\documentclass[letterpaper, 10pt, conference]{ieeeconf}  
\usepackage{amsmath,amsfonts}
\usepackage{algorithmic}
\usepackage{array}
\usepackage{textcomp}
\usepackage{stfloats}
\usepackage{url}
\usepackage{multirow}
\usepackage{arydshln}
\usepackage{verbatim}
\usepackage{graphicx}
\usepackage{subcaption}
\usepackage{booktabs} 
\usepackage{multirow} \usepackage{threeparttable} \usepackage{makecell} 
\usepackage{xcolor}
\usepackage{graphicx}
\usepackage{balance}
\usepackage{multirow}
\usepackage[ruled,vlined,linesnumbered]{algorithm2e}
\usepackage{xcolor}
\usepackage{hhline}    
\usepackage{color} 
\usepackage{bm}
\usepackage{amsthm}  
\usepackage{mathtools}
\usepackage{flushend}
\usepackage{xcolor}

\IEEEoverridecommandlockouts                              

\title{\LARGE \bf
Zero-Shot Signal Temporal Logic Planning with \par Disjunctive Branch Selection in Dynamic Semantic Maps
}

\author{Bowen~Ye,
	Ancheng~Hou, Junyue~Huang, Ruijia~Liu, and
	Xiang~Yin
	\thanks{This work was supported by the National Natural Science Foundation of China (62061136004, 62173226, 61803259). 
 }
\thanks{B. Ye, A. Hou, J. Huang, R. Liu and X. Yin  are with the School of Automation and Intelligent Sensing, Shanghai Jiao Tong University, and the Key Laboratory of System Control and Information Processing, the Ministry of Education of China, Shanghai 200240, China. {\tt  E-mail: \{yebowen1025,hou.ancheng,hjy-564904993,liuruijia, \
yinxiang\}@sjtu.edu.cn}.}
\thanks{
(Corresponding Author: Xiang Yin)
}
}

\def \until{{\mathbf{U}}}

\newtheorem{myrem}{Remark}

\begin{document}

\maketitle
 



\begin{abstract}
Signal Temporal Logic (STL) offers verifiable task specifications and is crucial for safety-critical control. Yet STL planning remains challenging: exact optimization-based methods are often too slow, and learning-based methods struggle to generalize across varying environments. We propose a zero-shot STL planning solver for variable-map environments that generates feasible trajectories without retraining. By integrating a map-conditioned Transformer architecture with a lightweight heuristic, our approach effectively handles complex disjunctive (OR) subformulas. Furthermore, we leverage Transitive Reinforcement Learning (TRL) to ensure consistent temporal grounding and logical coherence across decomposed sub-tasks. Experiments on dynamic semantic maps with diverse obstacle layouts demonstrate consistent gains, highlighting the framework's superior zero-shot generalization to changing environments and broad STL coverage.
\end{abstract}


\section{Introduction}
Formal methods~\cite{yin2024formal, belta2019formal} have become an essential tool for designing and deploying control systems in safety-critical settings, where failures can lead to severe consequences and where certification often requires explicit, machine-checkable guarantees. In such scenarios, it is not sufficient to specify goals in an informal or purely data-driven manner; instead, the task should be defined with precise semantics so that correctness can be rigorously analyzed, monitored, and enforced throughout execution. Among formal specification languages, Signal Temporal Logic (STL) has emerged as a particularly practical choice due to its ability to express rich temporal requirements over real-valued signals, making it well-suited for continuous-time or sampled-data dynamical systems and their sensor-driven observations. STL has therefore been widely adopted in robotics and control for mission specification, verification, and planning, enabling the integration of logical correctness constraints with quantitative robustness measures~\cite{sadraddini2015robust}.

Solving STL-constrained planning problems has been studied from multiple perspectives, yet each line of work exhibits some limitations when confronted with long-horizon, compositional STL specifications in changing environments. 
A classical approach encodes STL satisfaction into optimization, typically via mixed-integer formulations coupled with trajectory optimization~\cite{kurtz2022mixed,raman2014model}. 
While such methods offer strong rigor, their worst-case complexity grows rapidly with horizon length, map/obstacle complexity, and richer STL constructs often making them too slow for real-time deployment. 
To reduce computational burden, smooth or differentiable relaxations of STL robustness have been proposed to enable gradient-based optimization~\cite{gilpin2020smooth,leung2023backpropagation}; however, these objectives remain highly nonconvex, are sensitive to initialization, and still require per-instance iterative optimization at test time, which becomes brittle and costly when the map changes. 
In parallel, end-to-end learning-based planners attempt to amortize computation by mapping observations and STL specifications directly to actions or trajectories~\cite{meng2023signal}. 
Despite improved inference speed, they often struggle to learn and generalize the \emph{structured} and \emph{compositional} nature of STL (e.g., temporal nesting and disjunction), leading to unstable satisfaction and degraded performance outside the training distribution, particularly under variable-map settings. 
Reinforcement learning provides another route by treating STL robustness as a reward signal~\cite{aksaray2016q,kalagarla2021model}, but it inherits similar difficulties: long-horizon temporal constraints yield sparse and delayed rewards, exploration can violate safety constraints, and learning policies that consistently respect STL structure (especially disjunction) remains sample-inefficient and fragile under distribution shifts. 
Even in general trajectory generation, offline RL alleviates unsafe exploration by learning from logged data~\cite{kumar2020conservative}, yet its performance hinges on dataset coverage and it does not, on its own, guarantee reliable satisfaction of complex, structured STL specifications in unseen and changing maps.

Motivated by these considerations, recent research has pivoted towards zero-shot STL planning by decoupling high-level logical reasoning from low-level trajectory synthesis. 
This representative paradigm typically decomposes complex specifications into simpler, manageable subgoals, which are then realized by downstream modules such as trajectory generators or reactive policies~\cite{liu2025zero, meng2025mengtgpo, pmlr-v267-meng25b}. 
For instance, the  work by Liu \textit{et al.}~\cite{liu2025zero} established a robust foundation for this approach by leveraging a diffusion-based generator guided by parsed STL substructures, demonstrating impressive zero-shot capabilities within structured environments. 
In parallel, Meng \textit{et al.}~\cite{meng2025mengtgpo} introduced Temporal Grounded Policy Optimization (TGPO) to satisfy STL constraints through learned robustness signals. 
While these methods have significantly advanced the field, current reinforcement learning-based approaches~\cite{pmlr-v267-meng25b} primarily focus on single-layer STL tasks in relatively stable settings, leaving the extension to multi-layered logic and highly dynamic environments as a compelling next step.

Despite these promising strides, ensuring consistent generalization across \emph{variable-map} geometries remains an open challenge, as many existing pipelines are often tailored to specific scene representations or fixed environmental groundings. 
Furthermore, the handling of \emph{disjunctive} (OR) subformulas presents additional complexity; traditional decomposition may lead to sensitive branching decisions that are difficult for downstream modules to maintain under significant distribution shifts. 
These observations motivate the need for a more versatile zero-shot STL planner—one that can adapt to diverse map layouts while efficiently resolving complex disjunctive requirements.

To address the above challenges, we propose a \textbf{Transformer-based zero-shot STL solver} equipped with a \textbf{heuristic disjunction selection} mechanism for \emph{variable-map} environments. 
Following the ``decompose-then-synthesize'' paradigm, our solver first decomposes an STL specification into temporally grounded subgoals, and then generates a feasible trajectory to satisfy them. 
Unlike prior pipelines that rely on fixed scene grounding, our trajectory generator is a Transformer that \emph{explicitly fuses map information} with the task specification and state history, enabling robust planning under changing obstacle layouts and map geometries. 
In addition, we improve the \emph{time predictor} within the solver by training it with \emph{transitive reinforcement learning}~\cite{myers2025offline, park2025transitive}, leading to more accurate temporal grounding and consistently better overall performance.

Our main contributions are summarized as follows:
\begin{itemize}
    \item \textbf{Heuristic disjunction handling.} We introduce a lightweight yet effective heuristic for selecting and resolving disjunctive (OR) subformulas, substantially expanding the range of STL specifications that can be handled in practice.
    \item \textbf{Zero-shot planning on variable maps.} We design a map-conditioned Transformer trajectory generator that generalizes without retraining to unseen maps with different obstacle configurations and sizes, improving the universality of STL solving.
    \item \textbf{Strong performance across diverse environments.} Extensive experiments across environments with varying map complexity and obstacle layouts demonstrate strong and consistent performance, validating robustness and generalization in dynamic semantic maps.
\end{itemize}

\section{Related Works}
\subsection{Zero-shot Planner for STL}
Recent studies have begun to investigate \emph{zero-shot} STL planning, aiming to satisfy previously unseen STL specifications without retraining for each new task instance. 
A common recipe is to decouple logical reasoning from low-level control: the STL formula is first decomposed into simpler, temporally grounded subgoals, and a downstream module then realizes these subgoals through trajectory synthesis or policy execution~\cite{liu2025zero,meng2025mengtgpo}. 
Liu \textit{et al.}~\cite{liu2025zero} follow a generation-based approach, where parsed STL substructures guide a diffusion model to sample trajectories that meet the decomposed requirements. 
In contrast, Meng \textit{et al.}~\cite{meng2025mengtgpo} propose Temporal Grounded Policy Optimization (TGPO), which learns a temporally grounded policy using STL robustness-derived learning signals to improve satisfaction over time. 
These works demonstrate promising zero-shot capability across diverse task specifications; however, they often rely on environment-specific grounding or assume a fixed scene/map representation, which can limit generalization when map geometry and obstacle layouts change at test time.

\subsection{Transitive Reinforcement Learning}
Transitive reinforcement learning (TRL) has recently emerged as a principled framework for learning \emph{consistent} preference/ordering relations by exploiting transitivity across comparisons, providing an efficient way to infer latent structure from sparse supervision~\cite{myers2025offline,park2025transitive}. 
Rather than treating each pairwise signal independently, TRL propagates information through transitive constraints, which can improve sample efficiency and stabilize learning when the underlying objective admits a coherent ordering. 
In our setting, we leverage TRL to train a \emph{time predictor} that estimates the temporal allocation between two keypoints (or subgoals) produced by STL decomposition. 
By learning temporally consistent pairwise relations, the TRL-trained predictor provides more reliable temporal grounding for downstream trajectory synthesis, which in turn improves overall STL satisfaction and robustness in variable-map environments.

\subsection{Transformer-based Trajectory Generation}
Transformers~\cite{vaswani2017attention} have become a dominant backbone for trajectory generation and decision-making due to their strong sequence modeling capacity and flexible conditioning on heterogeneous context.
In reinforcement learning and planning, sequence-modeling formulations such as Decision Transformer and Trajectory Transformer cast control as conditional trajectory generation, enabling long-horizon reasoning with autoregressive decoding and search/planning over predicted sequences~\cite{chen2021decision,janner2021offline}.
In autonomous driving and robotics, Transformer architectures are also widely used to generate or forecast motion conditioned on map elements and multi-agent interactions, e.g., through scene-level attention over road geometry, agents, and time~\cite{nayakanti2022wayformer}.
More recently, tokenized motion generation further highlights the suitability of decoder-only Transformers for scalable trajectory synthesis in complex traffic scenes~\cite{wu2024smart}.
These advances motivate our design of a map-conditioned Transformer trajectory generator that fuses dynamic map information with task decomposition outputs, supporting efficient generation in variable-map environments.

\section{Preliminary}
In this section, we introduce some preliminaries for future problem formulation and solving procedure.
\subsection{Signal Temporal Logic Specification}

Let $X$ denote the state space and $\mathbf{x}_{0:T}=(x_0,\ldots,x_T)\in X^{T}$ be a finite trajectory.
The syntax of STL is
\begin{equation}
\phi ::= \top \mid \pi^{\mu} \mid \neg\phi \mid \phi_1 \wedge \phi_2 \mid \phi_1 \until_{[a,b]} \phi_2 ,
\end{equation}
where $\top$ denotes the Boolean constant \emph{true} and $\pi^{\mu}$ is an atomic predicate induced by
$\mu:\mathbb{R}^n\!\to\!\mathbb{R}$, which holds at time $k$ if $\mu(x_k)\ge 0$.
Standard Boolean connectives ($\neg,\wedge$) also induce $\vee$ and $\rightarrow$.
The temporal operator $\until_{[a,b]}$ (with $a,b\in\mathbb{R}_{\ge 0}$) is interpreted over state sequences:
$(\mathbf{x},t)\models \phi_1 \until_{[a,b]} \phi_2$ iff there exists $t'\in[t+a,t+b]$ such that
$(\mathbf{x},t')\models \phi_2$ and for all $t''\in[t+a,t']$, $(\mathbf{x},t'')\models \phi_1$.
We further use the derived operators $\mathbf{F}_{[a,b]}\phi := \top \until_{[a,b]} \phi$ (``eventually'') and
$\mathbf{G}_{[a,b]}\phi := \neg \mathbf{F}_{[a,b]}\neg \phi$ (``always'').
For brevity, we write $\mathbf{x}\models\phi$ as shorthand for $(\mathbf{x},0)\models\phi$.

Following prior work, we assume STL specifications are expressed in \emph{positive normal form} (PNF),
where negations appear only in front of atomic predicates, which facilitates subsequent reasoning and optimization~\cite{sadraddini2015robust}.

\subsection{DNF conversion and disjunction elimination.}
Given an STL task $\phi$, we convert it into a disjunctive normal form (DNF)
\begin{equation}
\tilde{\phi} \;=\; \phi_1 \,\vee\, \phi_2 \,\vee\, \cdots \,\vee\, \phi_n,
\end{equation}
where each subformula $\phi_i$ contains no explicit disjunction operators.
This conversion is obtained by recursively applying the following rewrite rules:
\begin{itemize}
  \item[(i)] $\mathbf{F}_{[a,b]}(\varphi_1 \vee \varphi_2) \;\Rightarrow\;
  \mathbf{F}_{[a,b]}\varphi_1 \,\vee\, \mathbf{F}_{[a,b]}\varphi_2$;
  \item[(ii)] $\mathbf{G}_{[a,b]}(\varphi_1 \vee \varphi_2) \;\Rightarrow\;
  \mathbf{G}_{[a,b]}\varphi_1 \,\vee\, \mathbf{G}_{[a,b]}\varphi_2$; and
  \item[(iii)] $(\varphi_1 \vee \varphi_2)\until_{[a,b]}(\psi_1 \vee \psi_2) \;\Rightarrow\;
  \displaystyle \bigvee_{i,j\in\{1,2\}} \; \varphi_i \until_{[a,b]} \psi_j$.
\end{itemize}
Importantly, the rewrites in (ii)--(iii) are not logically equivalent in general, and thus the resulting DNF
$\tilde{\phi}$ is a \emph{strengthening} of the original specification:
$\tilde{\phi}\Rightarrow \phi$ but not conversely.
This introduces conservativeness while preserving soundness, and similar simplifications are commonly adopted in STL planning and control synthesis to enable efficient downstream solving~\cite{yu2024continuous,liu2025zero}.

\subsection{Planning with Unknown Dynamics}\label{sec:unknown_dynamics}

In STL planning, the goal is to synthesize an action sequence such that the induced state (or output) signal satisfies a given STL specification.
When system dynamics are known, this can be addressed using model-based optimization and mixed-integer or differentiable formulations (e.g.,~\cite{kurtz2022mixed,raman2014model,gilpin2020smooth}).
In contrast, we consider a setting with \emph{unknown dynamics}.

\noindent\textbf{Unknown system and offline trajectories.}
We assume a discrete-time system
\begin{equation}
x_{t+1} = f(x_t,u_t), \qquad x_t\in\mathbb{R}^n,\ u_t\in\mathbb{R}^m,
\end{equation}
where the transition function $f:\mathbb{R}^n\times\mathbb{R}^m\to\mathbb{R}^n$ is unknown.
Instead, we are given an offline dataset $\mathcal{D}=\{\tau^{(i)}\}_{i=1}^{N}$ of historical operational trajectories,
collected from prior \emph{task-agnostic} executions that are consistent with the unknown dynamics.
Each trajectory $\tau^{(i)}=\{(x^{(i)}_t,u^{(i)}_t)\}_{t=0}^{T_i}$ may have a different length $T_i$.

\noindent\textbf{Dynamic map representation.}
Each planning instance is associated with a \emph{dynamic map} represented by a binary maze
$M \in \{0,1\}^{H\times W}$, where $M_{ij}=1$ indicates an occupied (obstacle) cell and $M_{ij}=0$ indicates free space.
The maze is defined on a grid with resolution $\texttt{cell\_size}>0$ (in meters), which induces a mapping between
continuous positions $p=(x,y)\in\mathbb{R}^2$ and grid indices $(i,j)\in\{0,\ldots,H\!-\!1\}\times\{0,\ldots,W\!-\!1\}$:
\begin{equation}
j = \Big\lfloor \frac{x}{\texttt{cell\_size}} \Big\rfloor,\qquad
i = \Big\lfloor \frac{y}{\texttt{cell\_size}} \Big\rfloor,
\end{equation}
and a position is considered collision-free if the corresponding cell is free, i.e., $M_{ij}=0$.
We allow $M$ (and thus obstacle layouts) to vary across test instances, capturing \emph{variable-map} environments.

\section{Problem Formulation}
\label{sec:problem}

We formalize the \emph{zero-shot STL planning} problem considered in this work.

\subsection{Planning Instance (Variable-Map Setting)}
A planning instance is specified by the tuple
\begin{equation}
\mathcal{I} \;=\; (M,\, x_0,\, \varphi),
\end{equation}
where $M \in \{0,1\}^{H\times W}$ is a binary maze map, $x_0$ is the initial state, and
$\varphi$ is an STL specification over the induced state signal.
We consider a \emph{variable-map} regime: across instances, the map $M$ may vary in \emph{layout, scale, and obstacle configuration}
(e.g., different obstacle placements and map sizes), while within an instance the map is fixed over the planning horizon.
Moreover, $\varphi$ is drawn from a \emph{full STL} fragment that may include disjunctions ($\vee$), thereby inducing branching satisfaction structure.

We assume access to an offline dataset $\mathcal{D}$ of trajectories consistent with the unknown system dynamics,
but no explicit dynamics model $f$ nor dynamics identifiers/parameters are available to the planner at inference time.

\subsection{Synthesis Objective and Feasibility Requirements}

Given an instance $\mathcal{I}$, the objective is to synthesize a finite-horizon trajectory
\begin{equation}
\mathbf{x}_{0:T} \;=\; (x_0,\ldots,x_T)
\end{equation}
that satisfies the STL specification with positive robustness while remaining collision-free and executable:
\begin{align}
\textbf{(STL satisfaction)}\quad
& \mathbf{x}_{0:T}\models \varphi, \label{eq:stl_pos} \\
\textbf{(Collision avoidance on }M\textbf{)}\quad
& M\!\left(\mathrm{idx}(p_t)\right)=0,\ \notag \\ \forall t\in\{0,\ldots,T\}, \label{eq:maze_free} \\
\textbf{(Execution feasibility)}\quad
& \mathbf{x}_{0:T}\in\mathcal{X}_{\mathrm{dyn}}(x_0), \label{eq:feasible_unknown}
\end{align}
where $p_t=(x_t,y_t)$ denotes the planar position component of $x_t$, $\mathrm{idx}(\cdot)$ is the position-to-grid mapping
defined in Sec.~II-C (via $\texttt{cell\_size}$ and floor discretization), and $\mathcal{X}_{\mathrm{dyn}}(x_0)$ denotes the set of trajectories realizable from $x_0$ under the true system dynamics.

Condition~\eqref{eq:feasible_unknown} formalizes the requirement that the synthesized trajectory must remain physically executable under the true dynamics.
In our evaluation, feasibility is assessed under distinct motion models (e.g., double-integrator and unicycle), while the planner itself is not provided with model identifiers or parameters during inference.

\subsection{Zero-Shot Generalization Requirement}

A solver operates in a \emph{zero-shot} manner if it can handle unseen instances $\mathcal{I}$ without instance-specific adaptation.
In particular, inference does not permit per-instance retraining, environment-specific fine-tuning, or map-dependent calibration.
Thus, the solver must generalize across varying maps $M$ and diverse STL specifications $\varphi$, including those with disjunction-induced branching.

\subsection{Intrinsic Challenges}

The formulation above is challenging due to the following coupled factors:
\begin{enumerate}
    \item \textbf{Branching induced by disjunctions.}
    Disjunctive subformulas induce satisfaction branching, especially under temporal nesting.
Prior zero-shot approaches often resolve disjunctions via unguided branch selection, which may choose branches that are map-infeasible or dynamically difficult, degrading success and efficiency.
    \item \textbf{Robustness to environment variability.}
    Practical deployments entail changing obstacle layouts and workspace geometries; therefore, supporting variable-map planning is essential for reliable operation beyond fixed environments.
    \item \textbf{Unknown dynamics at inference time.}
    In the absence of an explicit dynamics model, classical model-based STL synthesis (e.g., MILP/MIB formulations or shooting methods)
    cannot be directly applied, whereas purely data-driven methods often struggle to preserve the logical structure needed for broad STL coverage.
\end{enumerate}

\subsection{Targets}

Our goal is to develop an efficient solver that, for each instance $\mathcal{I}$, synthesizes trajectories satisfying
\eqref{eq:stl_pos}--\eqref{eq:feasible_unknown} while maintaining robustness and generalization across variable maps and full STL specifications (including disjunctions). Moreover, we aim to sustain high success rates under environment variability and to resolve disjunctive branching in a computationally efficient manner.

\begin{figure*}[t]
    \centering
    \includegraphics[width=0.85\textwidth]{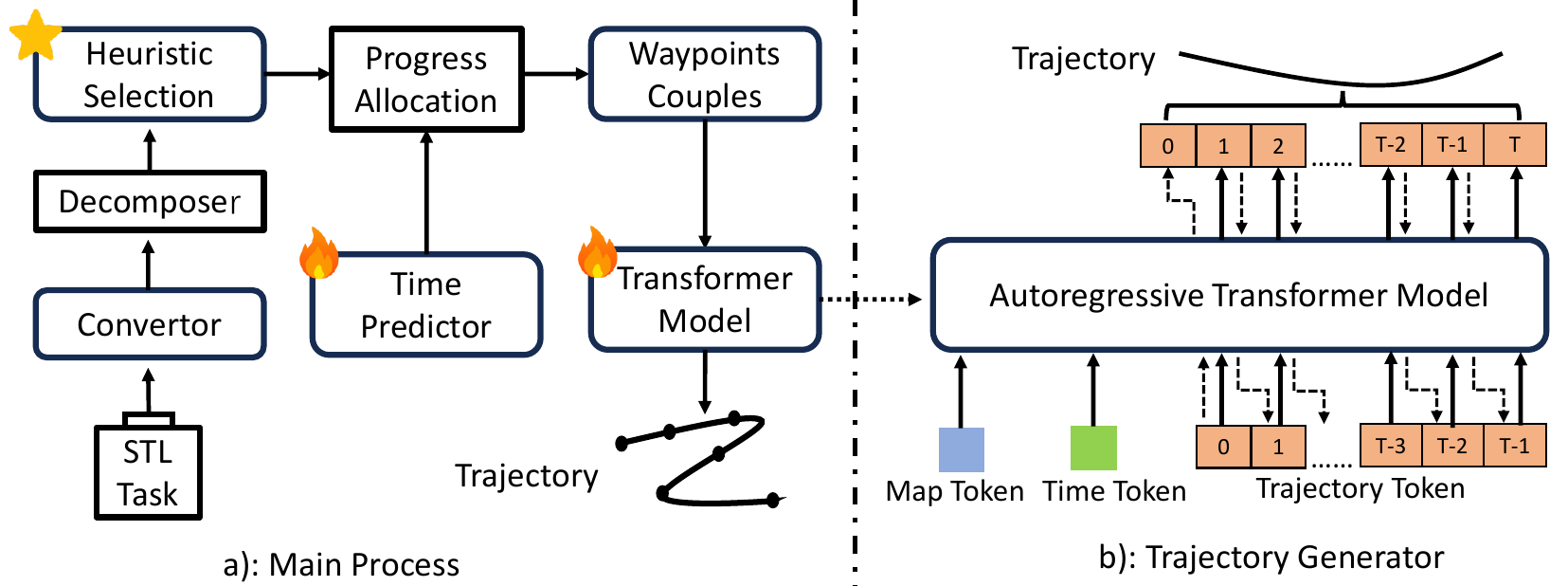}
    \caption{Overview of the STL-solving pipeline (left) and training of the autoregressive Transformer (right).}
    \label{fig:model}
    \vspace{-2em}
\end{figure*}

\section{Solving Procedure}

This section presents the proposed solver, which follows a decompose-then-synthesize paradigm. 
As shown in Fig.~\ref{fig:model}, the solver first resolves disjunctive branching in the STL formula, then grounds the resulting sub-tasks in time, and finally generates a complete trajectory using a map-conditioned autoregressive Transformer.

\subsection{Heuristic Selection of Branches}
\label{sec:heuristic_branch}

Disjunctive STL specifications are converted into a DNF-style form
$\tilde{\phi}=\bigvee_{k=1}^{K}\phi_k$, where each branch $\phi_k$ contains no explicit disjunction operators.
A practical limitation of prior zero-shot pipelines is that disjunctions are often resolved via unguided branch choice, which may select a branch that is infeasible in the given map or difficult to realize under unknown execution dynamics.
To address this issue, we introduce a lightweight heuristic that ranks candidate branches using three complementary criteria: (i) temporal-operator structure, (ii) map reachability, and (iii) temporal slack.
The branch with the minimal heuristic score is selected for downstream planning.

\medskip
\noindent\textbf{Inputs.}
The heuristic takes as input the binary maze map $M\in\{0,1\}^{H\times W}$ and the set of candidate branches
$\{\phi_k\}_{k=1}^{K}$.
Each branch can be parsed into a collection of temporal subformulas involving $\mathbf{F}$, $\mathbf{G}$, and $\mathbf{U}$ and their associated time intervals, as well as the set of region predicates referenced by the branch.

\medskip
\noindent\textbf{Temporal-operator complexity.}
We penalize branches with heavier temporal structure, reflecting the empirical difficulty of satisfying certain operators under long horizons and cluttered environments.
Let $N_F(\phi_k)$, $N_G(\phi_k)$, and $N_U(\phi_k)$ denote the numbers of $\mathbf{F}$-, $\mathbf{G}$-, and $\mathbf{U}$-type subformulas in $\phi_k$.
We define
\begin{equation}
C_{\mathrm{op}}(\phi_k)
\;=\;
w_F\,N_F(\phi_k) + w_G\,N_G(\phi_k) + w_U\,N_U(\phi_k),
\label{eq:hop}
\end{equation}
where $w_F<w_G<w_U$ (e.g., $w_F{=}10$, $w_G{=}20$, $w_U{=}30$).
This term is designed to dominate when the number of sub-tasks differs across branches, thereby favoring simpler decompositions.

\medskip
\noindent\textbf{Map reachability cost.}
We further incorporate a map-dependent term that discourages branches requiring regions that are difficult to access in the current maze.
We construct a connectivity graph over free cells in $M$ and compute grid-based reachability distances via graph search (e.g., BFS/DFS preprocessing).
Let $\mathcal{R}(\phi_k)$ denote the set of region predicates referenced by $\phi_k$, and let $d_M(R)\ge 0$ be the corresponding reachability cost for region $R$ under map $M$.
The map cost is defined as
\begin{equation}
C_{\mathrm{map}}(\phi_k)
\;=\;
\sum_{R\in\mathcal{R}(\phi_k)} d_M(R),
\label{eq:hmap}
\end{equation}
so that branches whose required regions are more readily reachable receive lower cost.

\medskip
\noindent\textbf{Temporal-slack reward.}
Finally, we prefer branches that admit larger temporal flexibility.
Let $\Delta(\phi_k)$ denote the set of time gaps extracted from $\phi_k$, including (i) the widths of the time intervals attached to subformulas and (ii) the separations between consecutive sub-tasks implied by the branch.
Define
\begin{equation}
C_{\mathrm{slack}}(\phi_k)
\;=\;
\sum_{\delta\in\Delta(\phi_k)} \sqrt{\delta},
\label{eq:hslack}
\end{equation}
which increases with available temporal slack and is therefore used as a negative term in the final score.

\medskip
\noindent\textbf{Heuristic score and selection.}
The overall heuristic score is computed as
\begin{equation}
S(\phi_k)
\;=\;
\alpha\,C_{\mathrm{op}}(\phi_k)
\;+\;
\beta\,C_{\mathrm{map}}(\phi_k)
\;-\;
\gamma\,C_{\mathrm{slack}}(\phi_k),
\label{eq:heuristic_score}
\end{equation}
where $\alpha,\beta,\gamma>0$ balance the three criteria.
In our implementation, we use $\alpha{=}0.2$, $\beta{=}0.2$, and $\gamma{=}0.1$.
We select the branch with minimum score,
\begin{equation}
k^\star \;=\; \arg\min_{k\in\{1,\ldots,K\}} S(\phi_k),
\qquad
\phi^\star \;=\; \phi_{k^\star},
\label{eq:select_branch}
\end{equation}
and pass $\phi^\star$ to the subsequent stages of the solver.

\begin{myrem}
    The proposed heuristic is intentionally lightweight: it requires only formula parsing and a one-time grid-based reachability preprocessing per map.
By discouraging operator-heavy branches, penalizing poorly reachable region requirements, and favoring larger temporal slack, it reduces the likelihood of selecting infeasible or unnecessarily difficult branches, thereby improving success rate and practical efficiency for disjunctive STL planning.
\end{myrem}

\subsection{Time Predictor}
\label{sec:time_predictor}

We employ a learned time predictor to provide temporal grounding for STL sub-tasks.
Given a map $M$, a start--goal query pair $(p_s,p_g)$ (in world coordinates), and $\texttt{cell\_size}$, the predictor outputs a continuous step estimate $\hat{\Delta t}\in\mathbb{R}_{\ge 0}$, which is rounded to an integer for downstream use.

\medskip
\noindent\textbf{Architecture.}
The predictor comprises (i) a CNN maze encoder producing a dense feature map and a global embedding, and (ii) an MLP regressor that fuses:
(a) locally sampled maze features at $p_s$ and $p_g$ (via differentiable bilinear sampling after world$\rightarrow$grid conversion), and
(b) geometric features of $(p_s,p_g)$ (e.g., offsets, distance, and Fourier encodings).
The final output is constrained to be non-negative (e.g., via a softplus activation).

\medskip
\noindent\textbf{Transitive Reinforcement Learning (TRL) objective.}
Beyond pointwise regression, we adopt \emph{Transitive Reinforcement Learning (TRL)} to improve temporal consistency.
Instead of only matching absolute step values, TRL enforces that predicted step lengths preserve \emph{relative orderings} implied by data, which provides a stronger supervisory signal and reduces sensitivity to scale and noise.
Concretely, within each minibatch we sample pairs of training examples $(i,j)$ with ground-truth step differences $\Delta t_i$ and $\Delta t_j$,
and impose a margin-based ranking constraint on the corresponding predictions $\hat{\Delta t}_i$ and $\hat{\Delta t}_j$.
Let $s_{ij}=\mathrm{sign}(\Delta t_i-\Delta t_j)\in\{-1,0,1\}$.
The TRL loss is defined as
\begin{equation}
\mathcal{L}_{\mathrm{TRL}}
=
\mathbb{E}_{(i,j)}
\Big[
\mathbb{I}[s_{ij}\neq 0]\,
\max\big(0,\; m - s_{ij}\,(\hat{\Delta t}_i-\hat{\Delta t}_j)\big)
\Big], \notag
\label{eq:trl}
\end{equation}
where $m>0$ is a margin.
This encourages $\hat{\Delta t}_i>\hat{\Delta t}_j$ whenever $\Delta t_i>\Delta t_j$, and vice versa.
Importantly, by training on many such pairwise comparisons, TRL promotes \emph{transitive} ordering consistency across multiple samples, improving global calibration of predicted time steps.

\medskip
\noindent\textbf{Training objective.}
The overall training loss combines pointwise accuracy with TRL regularization:
\begin{equation}
\mathcal{L}_{\text{time}}
=
\lVert \hat{\Delta t} - \Delta t \rVert_{1}
+
\lambda \,\mathcal{L}_{\mathrm{TRL}},
\label{eq:time_loss}
\end{equation}
where $\lambda$ controls the strength of the TRL term.

\medskip
\noindent\textbf{Inference.}
At test time, the predictor outputs $\hat{\Delta t}$ for each queried sub-task, and we use $\mathrm{round}(\hat{\Delta t})$ as the discrete time-step assignment for subsequent planning.

\subsection{Transformer-based Trajectory Generation}
\label{sec:traj_transformer}

Given the selected branch $\phi^\star$ and its associated temporal requirements, we generate a collision-free and executable trajectory using an autoregressive Transformer.
The model conditions on the environment map, obstacle/region descriptors, and the time-interval information extracted from the STL sub-tasks, and outputs a finite-horizon trajectory.

\medskip
\noindent\textbf{Inputs and tokenization.}
Let $M\in\{0,1\}^{H\times W}$ denote the binary maze map, and let $\mathcal{R}$ denote the set of region/obstacle descriptors referenced by $\phi^\star$ (e.g., target regions and avoid regions) together with their geometric extents on the grid.
Let $\mathcal{T}=\{[a_i,b_i]\}_{i=1}^{S}$ denote the collection of time intervals associated with the $S$ sub-tasks parsed from $\phi^\star$.
We form a conditioning set $\mathcal{C}=\{M,\mathcal{R},\mathcal{T}\}$ and embed it into a sequence of context tokens.
In particular, the map $M$ is encoded via a lightweight map encoder (e.g., patch embedding), region descriptors are encoded with a dedicated region encoder, and each time interval $[a_i,b_i]$ is mapped to a time embedding (e.g., through an MLP applied to $(a_i,b_i)$ and interval width).
These context tokens are concatenated with trajectory tokens representing the state sequence prefix.

\medskip
\noindent\textbf{Autoregressive Transformer decoder.}
We represent the output as a state trajectory $\mathbf{x}_{0:T}=(x_0,\ldots,x_T)$.
The generator is parameterized as an autoregressive conditional model:
\begin{equation}
p_\theta(\mathbf{x}_{1:T}\mid x_0,\mathcal{C})
\;=\;
\prod_{t=1}^{T} p_\theta\!\left(x_t \mid \mathbf{x}_{0:t-1}, \mathcal{C}\right),
\label{eq:ar_factorization}
\end{equation}
where $\theta$ denotes network parameters.
At each step $t$, the Transformer attends jointly to (i) the embedded context tokens derived from $\mathcal{C}$ and (ii) the embedded trajectory prefix $\mathbf{x}_{0:t-1}$, and outputs the next-state prediction $x_t$ (or a distribution over $x_t$).
Positional encodings are applied along the trajectory dimension to preserve temporal ordering.

\medskip
\noindent\textbf{Training with scheduled sampling.}
Teacher forcing can lead to exposure bias at inference.
To improve rollout robustness, we adopt scheduled sampling.
Let $\tilde{x}_{t-1}$ denote the token fed into the model at step $t$ as the prefix. We set
\begin{equation}
\tilde{x}_{t-1}
=
\begin{cases}
x_{t-1}^{\mathrm{gt}}, & \text{with probability } 1-\epsilon,\\[2pt]
x_{t-1}^{\mathrm{pred}}, & \text{with probability } \epsilon,
\end{cases}
\label{eq:scheduled_sampling}
\end{equation}
where $x_{t-1}^{\mathrm{gt}}$ is the ground-truth state and $x_{t-1}^{\mathrm{pred}}$ is the model prediction from the previous step.
The sampling rate $\epsilon$ is annealed over training (e.g., from $0$ to a target value), gradually transitioning from teacher forcing to fully autoregressive conditioning.

\begin{table*}[t]
  \caption{Experimental results in Maze2D.}
  \label{tab:results}
  \begin{small}
  \setlength{\tabcolsep}{10pt}
  \renewcommand{\arraystretch}{1.0}
  \begin{threeparttable}

    \begin{tabular*}{\linewidth}{@{\extracolsep{\fill}} c c cc cc cc @{}}
      \toprule
      \multirow{2}{*}{\textbf{Env}} & \multirow{2}{*}{\textbf{Dyn. Sys.}} &
      \multicolumn{2}{c}{\textbf{Success Rate(\%)$\uparrow$}} &
      \multicolumn{2}{c}{\textbf{Robustness Value$\uparrow$}} &
      \multicolumn{2}{c}{\textbf{Total Planning Time(s)$\downarrow$}} \\
      \cmidrule(lr){3-4}\cmidrule(lr){5-6}\cmidrule(lr){7-8}
      & & \textbf{baseline} & \textbf{ours}
        & \textbf{baseline} & \textbf{ours}
        & \textbf{baseline} & \textbf{ours} \\
      \midrule

      \multirow{2}{*}{U} & DI
        & $72.2$ & \textcolor{red}{$76.5$}
        & $1.4063$ & $1.3884$
        & $1.0013$ & $1.0323/\textcolor{blue}{0.3675}$ \\
      & UNI
        & \textcolor{red}{$50.4$} & $49.3$
        & $2.0490$ & $2.0295$
        & $1.0274$ & $1.1086/\textcolor{blue}{0.3896}$ \\
      \midrule

      \multirow{2}{*}{M} & DI
        & $57.0$ & \textcolor{red}{$63.9$}
        & $1.2768$ & $1.2531$
        & $1.0236$ & $1.0947/\textcolor{blue}{0.3547}$ \\
      & UNI
        & $47.5$ & \textcolor{red}{$51.7$}
        & $1.9536$ & $1.9555$
        & $1.1019$ & $1.1510/\textcolor{blue}{0.3923}$ \\
      \midrule

      \multirow{2}{*}{L} & DI
        & $48.0$ & \textcolor{red}{$54.8$}
        & $1.4229$ & $1.4838$
        & $1.0315$ & $1.0711/\textcolor{blue}{0.3345}$ \\
      & UNI
        & $42.5$ & \textcolor{red}{$47.8$}
        & $1.8257$ & $1.8960$
        & $1.0416$ & $1.0996/\textcolor{blue}{0.3472}$ \\
      \bottomrule
    \end{tabular*}

    \begin{tablenotes}[flushleft]
      \footnotesize
      \item Note: U/M/L denote $5{\times}5$, $7{\times}7$ and $9{\times}9$ maps, respectively. DI and UNI denote the double-integrator and unicycle dynamics. Results are reported as mean over $N$ seeds. \textit{Baseline} denotes our method without heuristic selection of disjunctive subformulas. The \textcolor{red}{red} denotes the best success rate. The \textcolor{blue}{blue} value after ``/'' reports the heuristic-selection time component.
    \end{tablenotes}

  \end{threeparttable}
  \end{small}
  \vspace{-2em}
\end{table*}

The model is optimized with a trajectory reconstruction objective, e.g., an $\ell_2$ loss on position and auxiliary state components, with optional feasibility regularizers consistent with the constraints in Sec.~\ref{sec:problem}.
At inference time, the model rolls out autoregressively by feeding back its predictions, producing the trajectory $\mathbf{x}_{0:T}$ conditioned on $\mathcal{C}$.

\subsection{STL Planner}
\label{sec:stl_planner}

Fig.~\ref{fig:model} illustrates the STL planning pipeline (left) and the training/inference mechanism of the trajectory generator (right).
Given an STL task, we first convert it into a disjunctive normal form and decompose it into a set of disjunction-free branches.
To mitigate the brittleness of unguided disjunction resolution, we apply the proposed heuristic selection (Sec.~\ref{sec:heuristic_branch}) to identify a branch $\phi^\star$ that is expected to be both map-feasible and easy to realize.

Conditioned on $\phi^\star$ and the map, we perform \emph{progress allocation} by grounding sub-tasks to discrete time locations.
This is implemented via the learned time predictor (Sec.~\ref{sec:time_predictor}), which estimates step distances between waypoint pairs implied by the sub-tasks, thereby producing a temporally consistent schedule over the planning horizon.
Finally, the autoregressive Transformer (Sec.~\ref{sec:traj_transformer}) synthesizes the full trajectory by jointly attending to map tokens, time tokens, and the trajectory prefix, yielding collision-free trajectories that satisfy the selected STL branch under the unknown execution dynamics.

Overall, the planner decouples logical branching, temporal grounding, and geometric realization, which enables scalable zero-shot planning across variable maps while improving robustness and efficiency when handling disjunctive STL specifications.

\section{Experiments}
We evaluate the proposed method in a grid-based navigation benchmark inspired by the U-Maze family of environments.
Each scene is represented as a binary maze map $M\in\{0,1\}^{H\times W}$, where $0$ denotes free space and $1$ denotes obstacles.
We consider three map sizes, namely $5{\times}5$, $7{\times}7$, and $9{\times}9$, to evaluate scalability with respect to environment size and complexity.
To bridge the discrete map representation and continuous-space planning, we embed the grid into $\mathbb{R}^2$ with a fixed resolution (\texttt{cell\_size}$=12$) and associate each grid index $(i,j)$ with its corresponding cell-center coordinate.
In addition, we assign a semantic label to each cell/region, while obstacle cells are marked with an explicit obstacle indicator to support predicate construction and collision checking.

For each map instance, we construct a suite of STL tasks by defining predicates over labeled regions (e.g., reach/avoid/stay constraints with specified temporal intervals).
Given a task $\phi$ and map $M$, our solver decomposes $\phi$ into temporally grounded subgoals and synthesizes a nominal continuous-space trajectory.
To assess dynamical feasibility, we execute the nominal plan with a standard trajectory-tracking controller to obtain a closed-loop rollout.
Unless stated otherwise, all metrics are computed on these executed trajectories, capturing both STL satisfaction and practical feasibility in variable-map settings.

All experiments are conducted on a workstation with an NVIDIA A800 GPU (80GB) and an AMD EPYC 7542 CPU (2 sockets, 32 cores per socket, 128 logical CPUs).
Training uses a single GPU, while evaluation and test-time rollouts are executed on CPU.

To evaluate generalization across dynamics, we consider two representative execution models: a double-integrator and a unicycle.
These systems impose distinct control and kinematic constraints, providing a complementary test of robustness.
Notably, the trajectory generator is not given an explicit dynamics model at inference time; it operates dynamics-agnostically and is evaluated solely through closed-loop rollouts under each execution system.

\subsection{Main Results}

\begin{figure*}[t]
    \centering

    \begin{subfigure}[t]{0.3\linewidth}
        \caption{\textbf{UNI (U scale)}}
        \includegraphics[width=\linewidth]{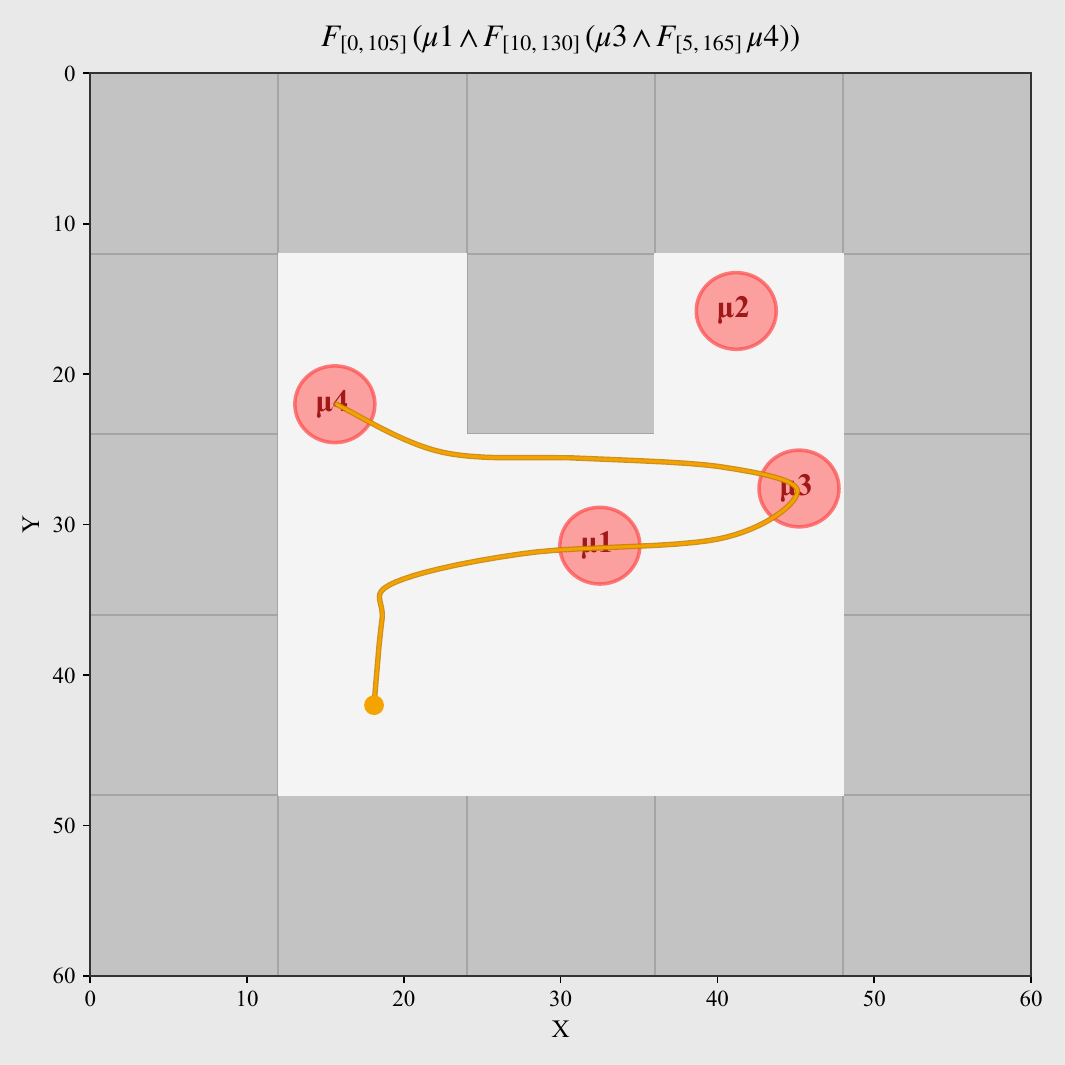}
    \end{subfigure}\hfill
    \begin{subfigure}[t]{0.3\linewidth}
        \caption{\textbf{UNI (M scale)}}
        \includegraphics[width=\linewidth]{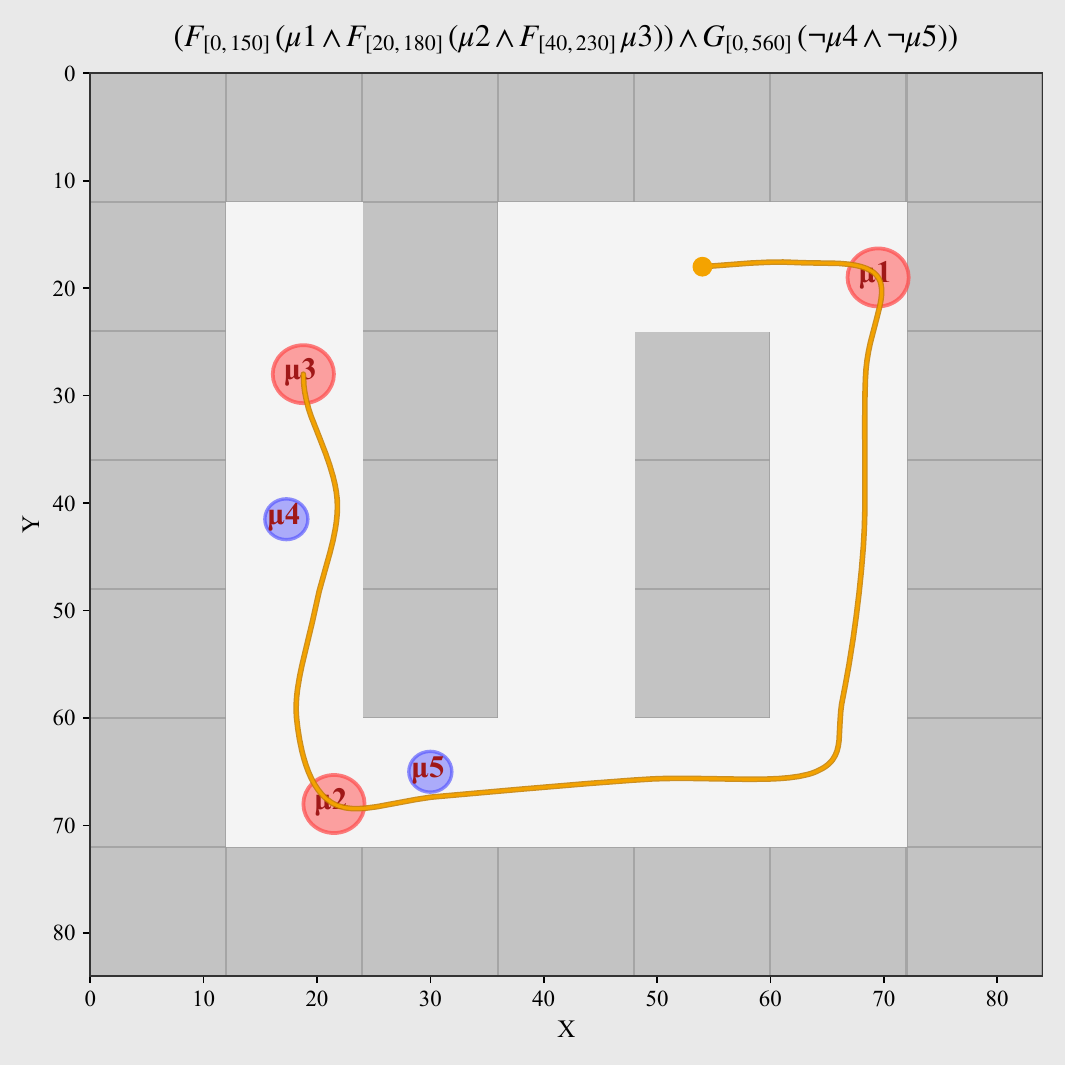}
    \end{subfigure}\hfill
    \begin{subfigure}[t]{0.3\linewidth}
        \caption{\textbf{UNI (L scale)}}
        \includegraphics[width=\linewidth]{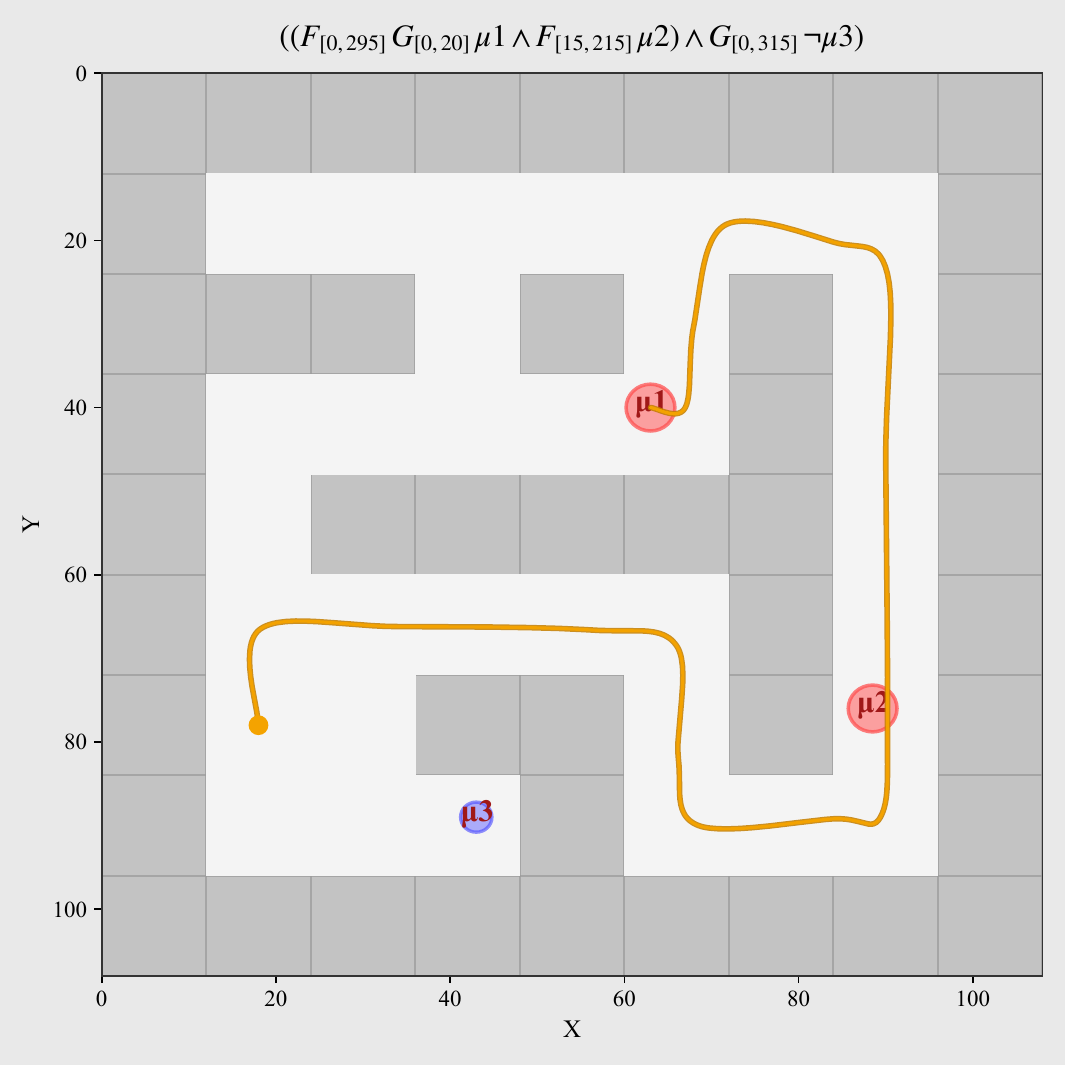}
    \end{subfigure}
    \begin{subfigure}[t]{0.3\linewidth}
        \caption{\textbf{DI (U scale)}}
        \includegraphics[width=\linewidth]{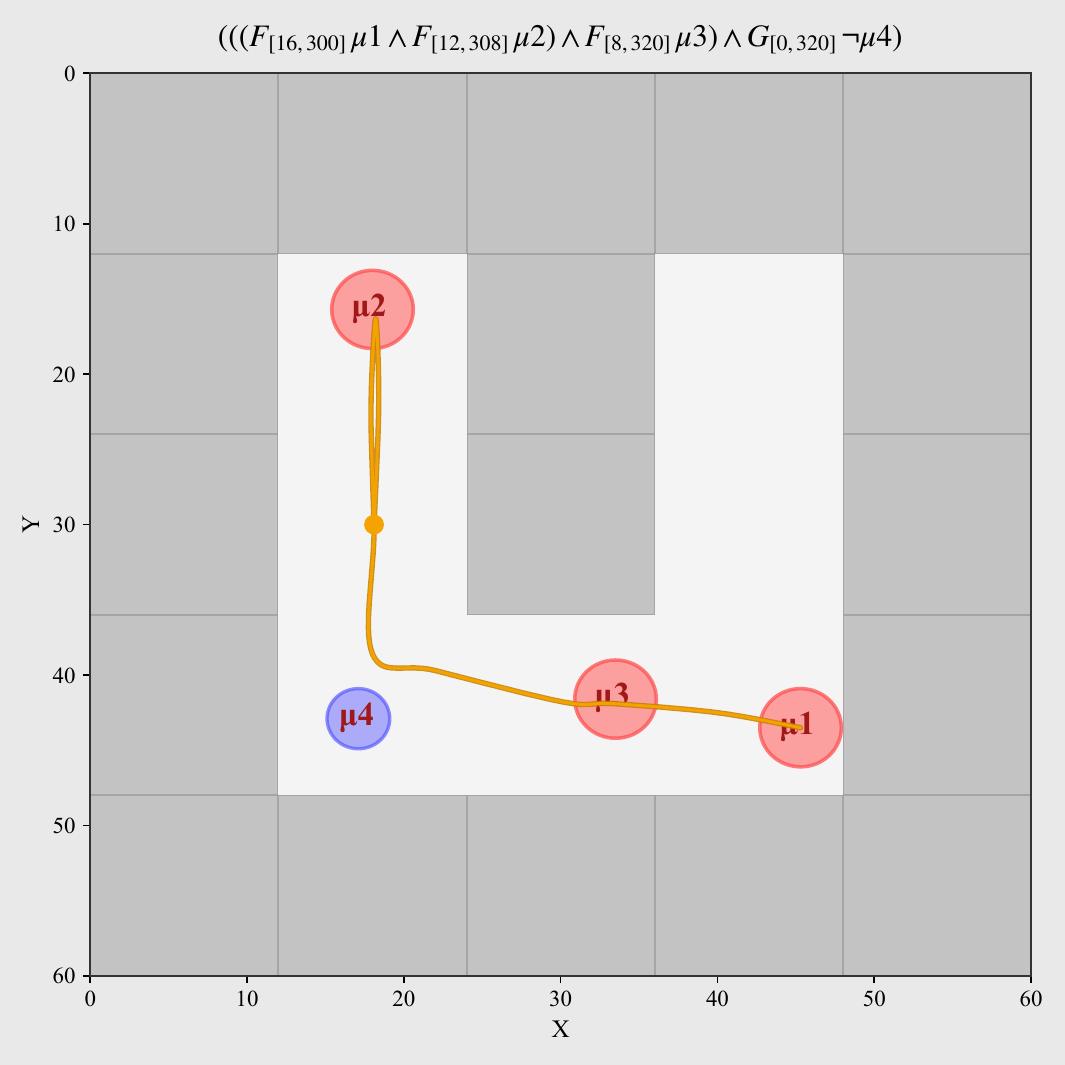}
    \end{subfigure}\hfill
    \begin{subfigure}[t]{0.3\linewidth}
        \caption{\textbf{DI (M scale)}}
        \includegraphics[width=\linewidth]{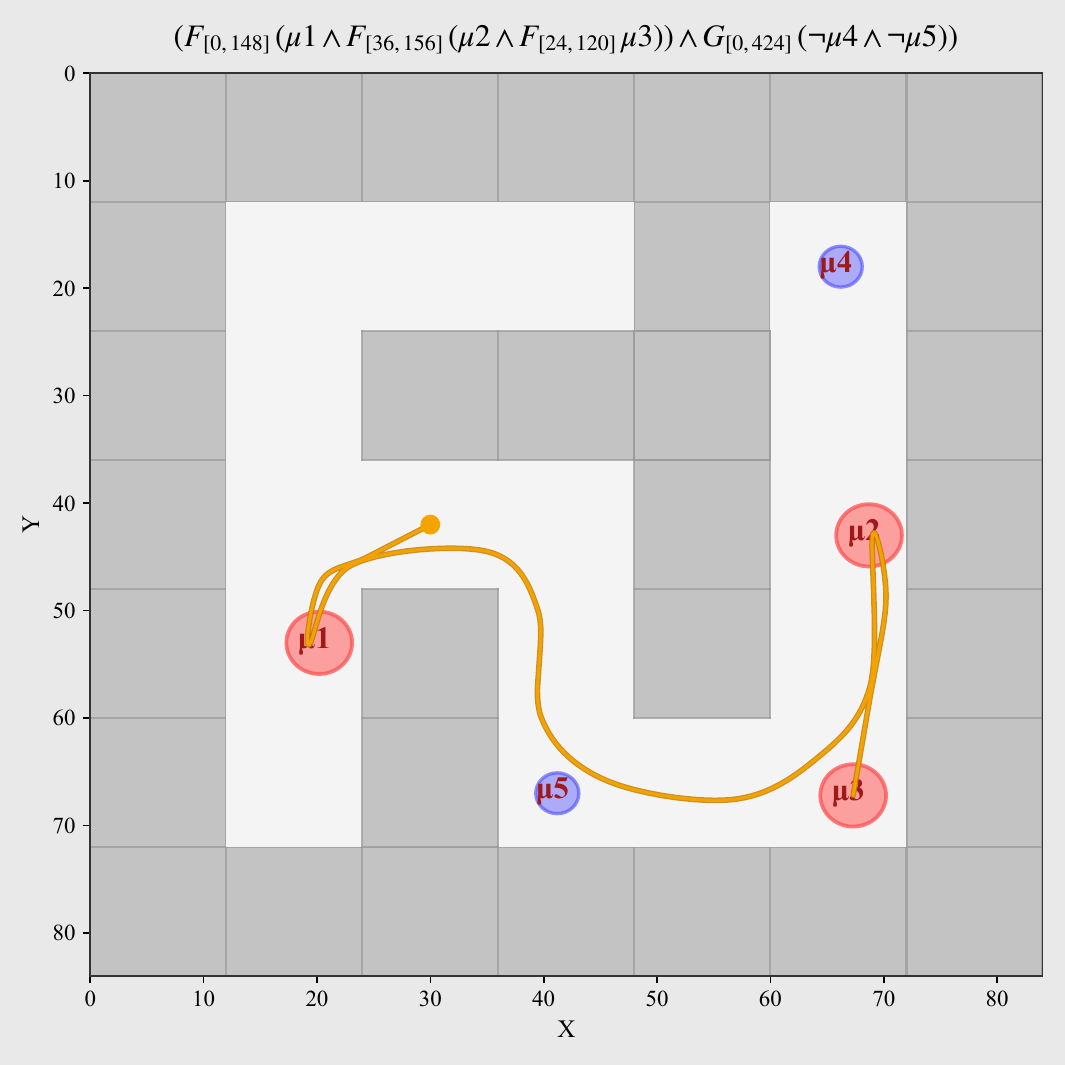}
    \end{subfigure}\hfill
    \begin{subfigure}[t]{0.3\linewidth}
        \caption{\textbf{DI (L scale)}}
        \includegraphics[width=\linewidth]{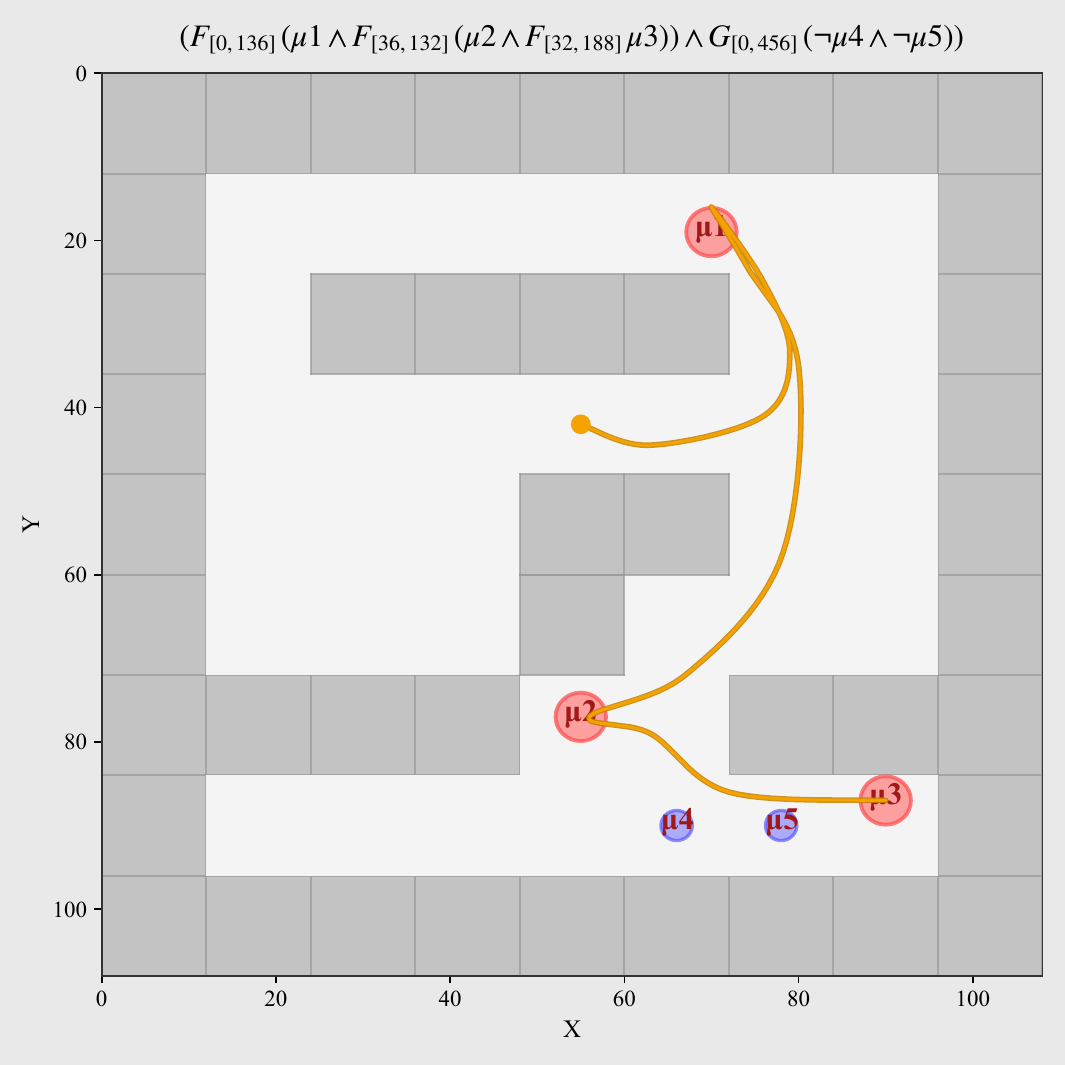}
    \end{subfigure}

    \caption{\textbf{Qualitative rollouts across dynamics and map scales.}
    Top row: unicycle (UNI). Bottom row: double-integrator (DI).
    For each subfigure, the STL specification after heuristic selection is shown at the top of the plot.}
    \label{fig:qual_six}
    \vspace{-2em}
\end{figure*}

Table~\ref{tab:results} summarizes performance across three map scales and two execution dynamics.
Overall, our method yields consistent success-rate improvements, with larger gains in more complex environments, indicating better scalability under increased constraint density.

Under double-integrator (DI) dynamics, we achieve higher success rates at all map sizes, with the most pronounced improvements in the medium and large settings.
This suggests that structure-aware subformula selection becomes more effective as the planning horizon grows and obstacle interactions intensify.
A similar trend holds under unicycle (UNI) dynamics, where gains are especially evident in the $7{\times}7$ and $9{\times}9$ cases.
These results support that the proposed decomposition mitigates the combinatorial branching induced by disjunctive STL specifications, particularly in long-horizon, cluttered scenes.
Importantly, robustness values remain comparable to, and in several cases exceed, the baseline, indicating that higher success rates are achieved without trading off specification margins.

In terms of efficiency, the nominal planning time is broadly comparable to the baseline, whereas enabling heuristic disjunctive selection substantially reduces the effective planning time (blue entries in Table~\ref{tab:results}).
The reduction grows with map size, suggesting that structure-aware pruning mitigates the computational overhead induced by temporal and logical branching.
This highlights the practical value of incorporating structured reasoning into STL planning pipelines.

Finally, the trajectory generator receives no explicit dynamics information at inference, yet it yields consistent improvements under both double-integrator and unicycle rollouts.
This supports the dynamics-agnostic property of the learned trajectory prior, implying that it captures geometry- and task-level structure that transfers across distinct motion models.

\subsection{Qualitative Experiment}
We present qualitative results to complement the quantitative evaluation in Table~\ref{tab:results}.
Figure~\ref{fig:qual_six} visualizes representative trajectories across three map scales (U/M/L), highlighting STL satisfaction and physical feasibility under different execution dynamics.

For each map scale, we report rollouts under two dynamics models, double-integrator (DI) and unicycle (UNI), with three examples per dynamics.
All trajectories satisfy the corresponding STL specification (shown at the top of each plot) while remaining collision-free and dynamically feasible.
For disjunctive specifications, the displayed formula is the \emph{selected} subformula after branch selection, i.e., the concrete task instance optimized by the planner.
These results suggest that the learned trajectory prior transfers across distinct motion models without requiring explicit dynamics information at inference time.

\begin{figure}[t]
    \centering
    \begin{subfigure}[t]{0.49\linewidth}
        \caption{\textbf{Heuristic selection (DI)}}
        \includegraphics[width=\linewidth]{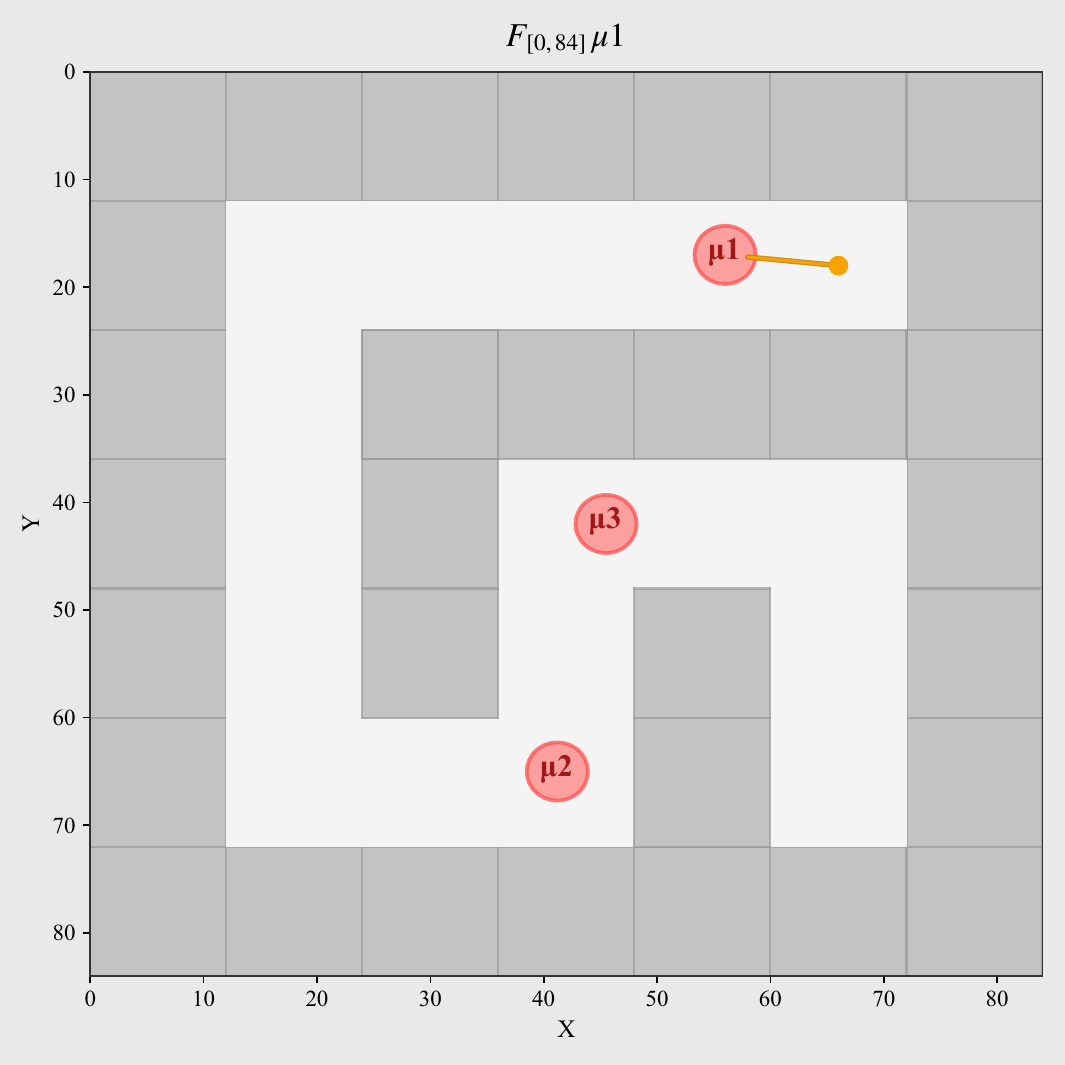}
    \end{subfigure}\hfill
    \begin{subfigure}[t]{0.49\linewidth}
        \caption{\textbf{Random selection (DI)}}
        \includegraphics[width=\linewidth]{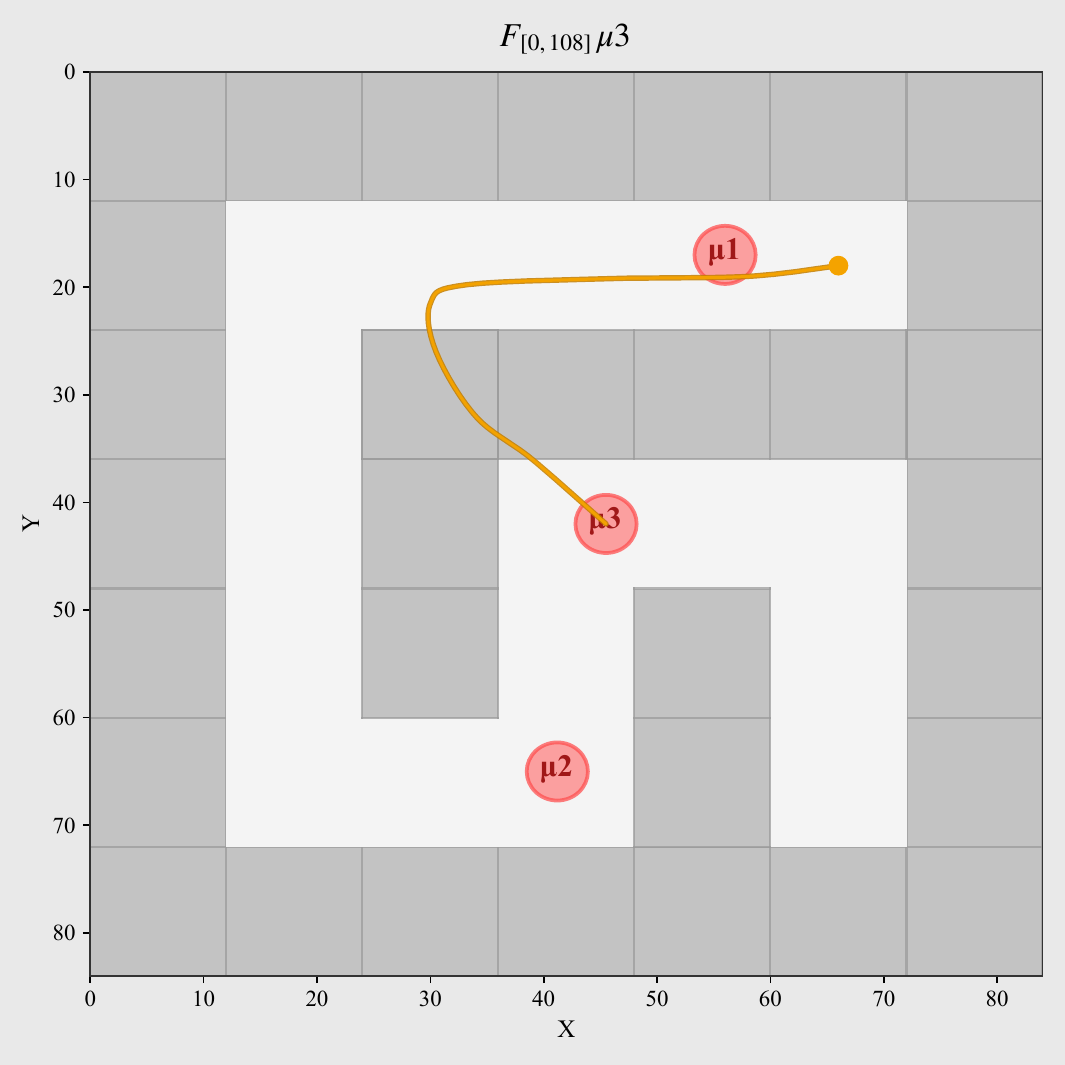}
    \end{subfigure}
    \caption{\textbf{Effect of disjunctive branch selection under DI dynamics.}
    Each plot shows the STL subformula after branch selection.
    Here, the disjunction is satisfied by reaching any one of three target regions.
    The heuristic preferentially selects a simpler, more feasible branch, whereas random selection may choose a harder branch and yield infeasible rollouts.}
    \label{fig:qual_heuristic}
    \vspace{-2em}
\end{figure}

To isolate the effect of disjunctive branch selection, Figure~\ref{fig:qual_heuristic} compares our structure-aware heuristic with a random-selection baseline under DI dynamics.
In this example, the disjunction permits reaching any of three target regions; the heuristic tends to select a simpler, more readily feasible branch, whereas random selection may choose a harder branch.
Consequently, the selected STL tasks differ between the two methods, and random selection can lead to infeasible rollouts.
Overall, these examples underscore the robustness of the generated behaviors and the practical benefit of structured pruning in disjunctive STL planning.

\section{Conclusions}\label{sec:conclusion}
In this work, we proposed a zero-shot STL planner for 
variable-map environments, addressing a key challenge 
in deploying STL-based planners to real-world scenarios 
where maps are not fixed at training time. By combining a map-conditioned Transformer trajectory generator with a decompose-then-synthesize pipeline, our method handles complex disjunctive STL formulas without requiring environment-specific retraining. The introduction of transitive reinforcement learning for time prediction further bridges the gap between symbolic temporal logic and learned trajectory planning.
Experiments across diverse dynamic semantic maps 
demonstrate consistent improvements in STL satisfaction, 
suggesting that structured decomposition and 
map-conditioning are promising directions for 
generalizable robot planning.

\bibliographystyle{plain}
\bibliography{myref}

@article{kurtz2022mixed,
  title={Mixed-integer programming for signal temporal logic with fewer binary variables},
  author={Kurtz, Vincent and Lin, Hai},
  journal={IEEE Control Systems Letters},
  volume={6},
  pages={2635--2640},
  year={2022},
  publisher={IEEE}
}

@inproceedings{aksaray2016q,
  title={Q-learning for robust satisfaction of signal temporal logic specifications},
  author={Aksaray, Derya and Jones, Austin and Kong, Zhaodan and Schwager, Mac and Belta, Calin},
  booktitle={55th IEEE  Conference on Decision and Control (CDC)},
  pages={6565--6570},
  year={2016}
}

@article{meng2023signal,
  title={Signal temporal logic neural predictive control},
  author={Meng, Yue and Fan, Chuchu},
  journal={IEEE Robotics and Automation Letters},
  volume={8},
  number={11},
  pages={7719--7726},
  year={2023}
}

@inproceedings{raman2014model,
  title={Model predictive control from signal temporal logic specifications: A case study},
  author={Raman, Vasumathi and Maasoumy, Mehdi and Donz{\'e}, Alexandre},
  booktitle={Proceedings of the 4th ACM SIGBED international workshop on design, modeling, and evaluation of cyber-physical systems},
  pages={52--55},
  year={2014}
}

@inproceedings{kalagarla2021model,
  title={Model-free reinforcement learning for optimal control of Markov decision processes under signal temporal logic specifications},
  author={Kalagarla, Krishna C and Jain, Rahul and Nuzzo, Pierluigi},
  booktitle={60th IEEE Conference on Decision and Control (CDC)},
  pages={2252--2257},
  year={2021}
}

@article{liu2025zero,
  title={Zero-Shot Trajectory Planning for Signal Temporal Logic Tasks},
  author={Liu, Ruijia and Hou, Ancheng and Yu, Xiao and Yin, Xiang},
  journal={arXiv:2501.13457},
  year={2025}
}

@article{myers2025offline,
  title={Offline goal-conditioned reinforcement learning with quasimetric representations},
  author={Myers, Vivek and Zheng, Bill Chunyuan and Eysenbach, Benjamin and Levine, Sergey},
  journal={arXiv:2509.20478},
  year={2025}
}

@article{park2025transitive,
  title={Transitive RL: Value Learning via Divide and Conquer},
  author={Park, Seohong and Oberai, Aditya and Atreya, Pranav and Levine, Sergey},
  journal={arXiv:2510.22512},
  year={2025}
}

@article{kumar2020conservative,
  title={Conservative q-learning for offline reinforcement learning},
  author={Kumar, Aviral and Zhou, Aurick and Tucker, George and Levine, Sergey},
  journal={Advances in neural information processing systems},
  volume={33},
  pages={1179--1191},
  year={2020}
}

@article{belta2019formal,
  title={Formal methods for control synthesis: An optimization perspective},
  author={Belta, Calin and Sadraddini, Sadra},
  journal={Annual Review of Control, Robotics, and Autonomous Systems},
  volume={2},
  number={1},
  pages={115--140},
  year={2019},
  publisher={Annual Reviews}
}

@article{yin2024formal,
  title={Formal synthesis of controllers for safety-critical autonomous systems: Developments and challenges},
  author={Yin, Xiang and Gao, Bingzhao and Yu, Xiao},
  journal={Annual Reviews in Control},
  volume={57},
  pages={100940},
  year={2024},
  publisher={Elsevier}
}

@article{gilpin2020smooth,
  title={A smooth robustness measure of signal temporal logic for symbolic control},
  author={Gilpin, Yann and Kurtz, Vince and Lin, Hai},
  journal={IEEE Control Systems Letters},
  volume={5},
  number={1},
  pages={241--246},
  year={2020},
  publisher={IEEE}
}

@article{leung2023backpropagation,
  title={Backpropagation through signal temporal logic specifications: Infusing logical structure into gradient-based methods},
  author={Leung, Karen and Ar{\'e}chiga, Nikos and Pavone, Marco},
  journal={The International Journal of Robotics Research},
  volume={42},
  number={6},
  pages={356--370},
  year={2023},
  publisher={SAGE Publications Sage UK: London, England}
}

@inproceedings{sadraddini2015robust,
  title={Robust temporal logic model predictive control},
  author={Sadraddini, Sadra and Belta, Calin},
  booktitle={53rd Annual Allerton Conference on Communication, Control, and Computing},
  pages={772--779},
  year={2015}
}

@article{yu2024continuous,
  title={Continuous-time control synthesis under nested signal temporal logic specifications},
  author={Yu, Pian and Tan, Xiao and Dimarogonas, Dimos V},
  journal={IEEE Transactions on Robotics},
  volume={40},
  pages={2272--2286},
  year={2024},
  publisher={IEEE}
}

@article{meng2025mengtgpo,
  title={{TGPO}: Temporal Grounded Policy Optimization for Signal Temporal Logic Tasks},
  author={Meng, Yue and Chen, Fei and Fan, Chuchu},
  journal={arXiv:2510.00225},
  year={2025}
}

@article{vaswani2017attention,
  title={Attention is all you need},
  author={Vaswani, Ashish and Shazeer, Noam and Parmar, Niki and Uszkoreit, Jakob and Jones, Llion and Gomez, Aidan N and Kaiser, {\L}ukasz and Polosukhin, Illia},
  journal={Advances in neural information processing systems},
  volume={30},
  year={2017}
}

@article{chen2021decision,
  title={Decision transformer: Reinforcement learning via sequence modeling},
  author={Chen, Lili and Lu, Kevin and Rajeswaran, Aravind and Lee, Kimin and Grover, Aditya and Laskin, Misha and Abbeel, Pieter and Srinivas, Aravind and Mordatch, Igor},
  journal={Advances in neural information processing systems},
  volume={34},
  pages={15084--15097},
  year={2021}
}

@article{janner2021offline,
  title={Offline reinforcement learning as one big sequence modeling problem},
  author={Janner, Michael and Li, Qiyang and Levine, Sergey},
  journal={Advances in neural information processing systems},
  volume={34},
  pages={1273--1286},
  year={2021}
}

@article{nayakanti2022wayformer,
  title={Wayformer: Motion forecasting via simple \& efficient attention networks},
  author={Nayakanti, Nigamaa and Al-Rfou, Rami and Zhou, Aurick and Goel, Kratarth and Refaat, Khaled S and Sapp, Benjamin},
  journal={arXiv preprint arXiv:2207.05844},
  year={2022}
}

@article{wu2024smart,
  title={Smart: Scalable multi-agent real-time motion generation via next-token prediction},
  author={Wu, Wei and Feng, Xiaoxin and Gao, Ziyan and Kan, Yuheng},
  journal={Advances in Neural Information Processing Systems},
  volume={37},
  pages={114048--114071},
  year={2024}
}

@InProceedings{pmlr-v267-meng25b,
  title = 	 {{T}e{L}o{G}ra{F}: Temporal Logic Planning via Graph-encoded Flow Matching},
  author =       {Meng, Yue and Fan, Chuchu},
  booktitle = 	 {Proceedings of the 42nd International Conference on Machine Learning},
  pages = 	 {43754--43780},
  year = 	 {2025}
}

\end{document}